# Advancing Green AI: Efficient and Accurate Lightweight CNNs for Rice Leaf Disease Identification

Khairun Saddami[a,b], Yudha Nurdin[a], Mutia Zahramita[a], Muhammad Shahreeza Safiruz[c]

[a]Department of Electrical and Computer Engineering, Faculty of Engineering, Universitas Syiah Kuala

[b]Telematic Research Centers, Universitas Syiah Kuala

[c]Department of Artificial Intelligence, Faculty of Computer Science and Information Technology, University of Malaya

## Abstract

Rice plays a vital role as a primary food source for over half of the world's population, and its production is critical for global food security. Nevertheless, rice cultivation is frequently affected by various diseases that can severely decrease yield and quality. Therefore, early and accurate detection of rice diseases is necessary to prevent their spread and minimize crop losses. In this research, we explore three mobile-compatible CNN architectures, namely ShuffleNet, MobileNetV2, and EfficientNet-B0, for rice leaf disease classification. These models are selected due to their compatibility with mobile devices, as they demand less computational power and memory compared to other CNN models. To enhance the performance of the three models, we added two fully connected layers separated by a dropout layer. We used early stop creation to prevent the model from being overfiting. The results of the study showed that the best performance was achieved by the EfficientNet-B0 model with an accuracy of 99.8%. Meanwhile, MobileNetV2 and ShuffleNet only achieved accuracies of 84.21% and 66.51%, respectively. This study shows that EfficientNet-B0 when combined with the proposed layer and early stop, can produce a high-accuracy model.

Keywords: rice leaf detection; green AI; smart agriculture; EfficientNet;

## Introduction

Rice is a staple food crop for more than half of the world's population, and its production is crucial for global food security [1]. As one of the most important crops globally, rice provides sustenance for billions of people, particularly in Asia and Africa. Its production and yield are vital to the agricultural industry and the global economy. However, rice cultivation is often threatened by various diseases that can significantly reduce yield and quality. Early and accurate detection of rice diseases is essential to prevent their spread and minimize crop losses. Detecting rice diseases in their early stages can help farmers take prompt action, such as applying appropriate fungicides or changing cultural practices. This can significantly reduce the spread of the disease and minimize crop losses. Moreover, accurate disease detection can help farmers optimize their crop management practices, such as irrigation, fertilization, and harvesting, to maximize yield and quality.

In recent years, machine learning and computer vision techniques have been applied to remote sensing data to develop automated rice disease detection systems [2]. These systems can analyze the images of the rice fields and identify the symptoms of the diseases, providing early and accurate detection of rice diseases. Moreover, these systems can be integrated with other technologies, such as IoT and cloud computing, to provide real-time monitoring and decision-making support for farmers [3,4]. However, building a real-time system that is based on IoT requires a low-cost and efficient computing system. This is where green AI comes in, which refers to artificial intelligence systems that have high accuracy but low computational requirements. Green AI is particularly suitable for real-time and IoT-based systems, as it can provide accurate predictions while minimizing energy consumption and computational costs.

A range of studies have identified Brown Spot, Healthy, Hispa, and Leaf Blast as the most common diseases in rice leaves [5-8]. These diseases can significantly impact rice production and food security, making their early detection and classification crucial. Various machine learning and deep learning techniques have been employed to detect and classify these diseases, with promising results accurately. For example, a 5-layer convolutional model achieved an accuracy of 78.2% [5], while a proposed CNN technique outperformed other models with an accuracy rate of 93% [6]. The use of fusion features and machine learning algorithms has also shown improved accuracy in disease identification [7]. Furthermore, a classification model based on leaf color using the MobileNet model achieved 97% accuracy [8]. These studies collectively highlight the potential of advanced technologies in effectively managing rice leaf diseases.

In recent years, mobile-based computer vision techniques have gained popularity for crop disease detection due to their low cost, ease of use, and potential for real-time monitoring [9-10]. Among these techniques, convolutional neural network (CNN) architectures have shown promising results in detecting and classifying rice diseases with high accuracy. In addition to these advantages, models built on mobile devices are lightweight. A lightweight model can lead to lower energy consumption and reduced computation, resulting in energy savings that can help prevent accelerated climate change.

In this study, we investigate the application of three mobile-based CNN architectures, namely ShuffleNet, MobileNetV2, and EfficientNet-B0, for the classification of rice diseases. These architectures are chosen because of their suitability for mobile devices, as they require less computational power and memory compared to other CNN models. The proposed method involves training the CNN models on a large dataset of rice disease images and evaluating their performance in terms of accuracy, precision, recall, and F1-score. The results show that the proposed method can accurately classify different rice diseases with high accuracy and robustness.

The study's contribution is significant, as it provides a practical and cost-effective solution for rice disease detection and monitoring. The proposed method can be easily integrated into existing mobile applications and can help farmers and agricultural extension workers to make timely and informed decisions about rice cultivation and management. The findings of this study can also be extended to other crops and diseases and can contribute to the development of more advanced and intelligent agricultural systems.

In the following sections, we describe the related works and present the material and method for conducting the experiment. Then, we present and discuss the results and provide recommendations for further research in the conclusion section.

## Related Works

In 2023, Simhadri's research published in the MDPI Agronomy journal used the Inceptionv3 model and achieved a 99% accuracy on a public dataset from Mendeley and Kaggle. While the accuracy is impressive, the model used is quite heavy and large, with 24 M parameters and a size of 89 MB. This model is relatively large for a deep learning model used for classification purposes. The large number of learning parameters will undoubtedly affect the computation process [11]. During that year, Hasan's research published in the MDPI Agriculture Journal produced a model with a classification accuracy of 97% [12]. Hasan built a CNN from scratch with 3.98 M parameters. The dataset used was from Kaggle, consisting of 2700 rice leaf images. However, the model only classified three classes:

bacterial leaf blast, leaf smut, and brown spot, without including a healthy rice leaf class. Including a healthy class would be significant for more generalized use. Additionally, while the research claims to be superior, it also includes additional steps that can increase computational complexity.

In 2022, Biplob Dey conducted a comparison of two types of rice leaf diseases caused by fungi [13]. The dataset used was a combination of public and private datasets. The research showed that VGG19 produced the highest accuracy, with a score of 92.4%. Although this accuracy is good, the model used is quite heavy and large. VGG19 has approximately 144 million parameters and a size of 535 MB, which is quite large for a deep learning model used for classification purposes. The large number of learning parameters will undoubtedly affect the computation process. In the same year, Agrawal also conducted a comparison study of several CNN architectures for detecting diseases in rice plants [14]. The dataset used was taken from various sources, with a total of 5000 datasets. The comparison resulted in ResNet50 having the highest accuracy, with a score of 97.5%. However, this architecture still needs to be suitable for implementation in mobile devices as the model is quite large and has many parameters.

In addition to rice plants, disease detection in plant leaves has also been conducted by several researchers, including those studying cassava [15-16], tomato plants [17-18], corn plants [19-20] and apples [21-22]. Research findings indicate that the focus is still on achieving high accuracy rather than on creating appropriate and implementable models for mobile devices. According to several articles, such as [23-29], transformer or ViT models achieve the highest accuracy. However, ViT architecture is quite large and has a very large model size. A large architecture size significantly affects the required computational resources. The smaller the size of an architecture, the smaller the required resources. Using fewer resources can lead to energy savings, which ultimately contributes to green AI.

The need for small or lightweight models is becoming increasingly important in the context of green AI systems, which aim to reduce the environmental impact of AI models. Large models, such as the ViT architecture, require significant computational resources, including processing power, memory, and energy. These resources are often provided by data centers, which consume a substantial amount of electricity and contribute to greenhouse gas emissions.

By developing smaller or lightweight models, we can reduce the computational resources required to train and deploy these models. This method can lead to a significant reduction in energy consumption and greenhouse gas emissions, contributing to the fight against climate change. Moreover, smaller models are more suitable for implementation on mobile devices, which can enable real-time, on-device processing and analysis of plant disease images. It can lead to faster and more accurate detection of plant diseases, which can help to prevent their spread and minimize crop losses. In summary, developing small or lightweight models for plant disease detection is essential for creating green AI systems that are both environmentally sustainable and practical for real-world applications. By reducing the computational resources required for plant disease detection, we can contribute to the fight against climate change and promote sustainable agriculture.

## Material and Method

In this section, we describe the material, including the dataset used and the method used to conduct the experiment.

### Dataset Preparation

The dataset used in this research was obtained from Kaggle[1]. This dataset consists of four classes of rice leaf diseases: Brown spot, Healthy, Hispa, and Leaf Blast. The dataset structure is organized into four folders, each containing images of rice plant leaves in "jpg" format. An example image of rice leaf diseases can be seen in Figure 1.

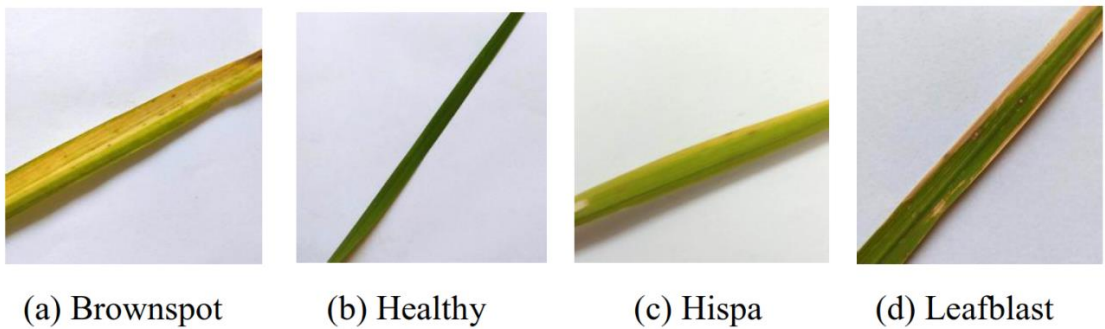

Figure 1 Example of rice plant leaf with disease

To ensure a balanced dataset for optimal deep learning model performance during training, 2,092 images were used from the original 3,355 rice plant disease images. Downsampling was performed so that the model had an equal number of images per class, with 523 images for each class: Brownspot, Healthy, Hispa, and Leaf Blast. To enhance the variation of the dataset, we augment the training dataset using six image augmentations, including rotation range with an angle of 30 degrees, zoom range with a scale of 0.15, width shift range with a value of 0.2, height shift range with value is 0.2, shear range with a value of 0.15, and horizontal_flip.

[1](https://www.kaggle.com/datasets/shayanriyaz/riceleafs)

## Experimental Method

To develop a classification model for rice leaf disease, we deployed three lightweight baseline models that are claimed to be suitable for mobile devices, e.g. EfficientNet-B0, MobileNetV2, and ShuffleNet. To enhance accuracy and prevent overfitting during the training process, we added a new block with additional layers, including a fully connected (FC) layer, a dropout layer, and another FC layer, as shown in Figure 2. The number of features used in the FC layer is 128 for each layer, while the dropout layer has a value of 0.3. This value was found to be the best during the experiment. Additionally, we added L2 regularization to prevent overfitting by setting the lambda to $10^{-4}$.

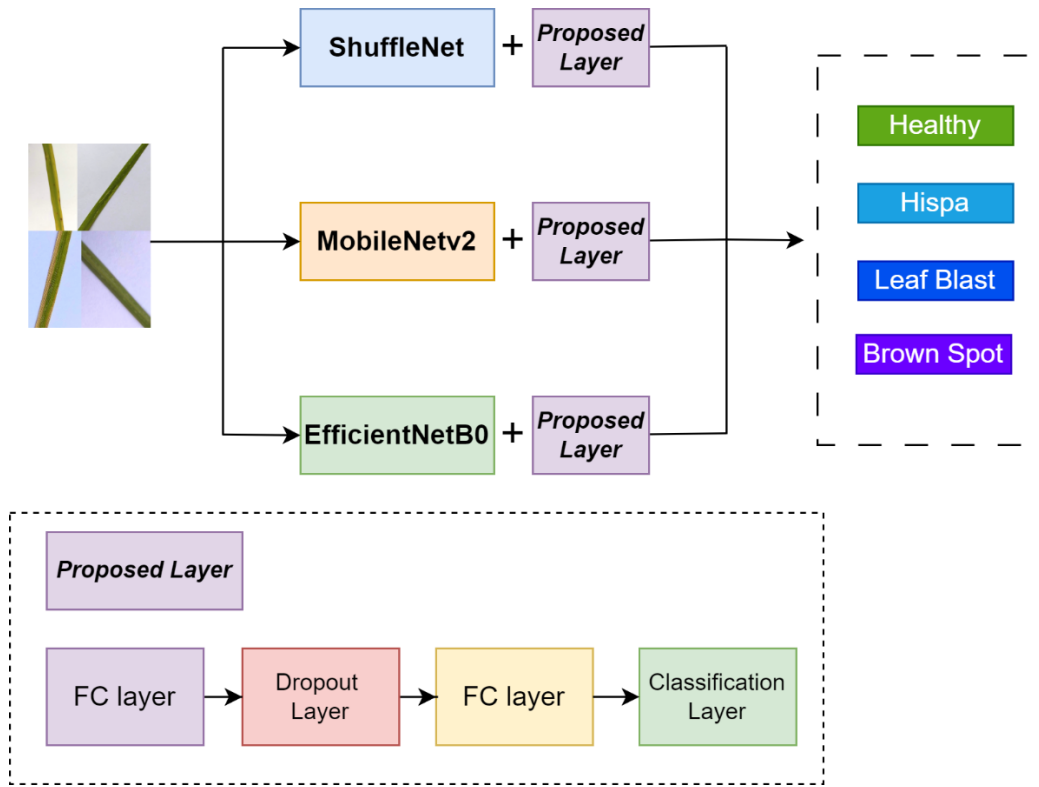

Figure 2. The proposed method

## CNN Pre-trained Model

EfficientNet is a family of convolutional neural network models known for their faster training speed and better parameter efficiency compared to previous models [30]. EfficientNet can enhance network performance, covering width, depth, and resolution. This model can achieve higher accuracy and better performance compared to other CNN architectures through compound scaling. Figure 3 presents the EfficientNet-B0 architecture algorithm, which uses the "compound scaling" approach, adjusting the width, depth, and image resolution simultaneously to optimize the model's performance across various computational levels. Additionally, this architecture takes advantage of depthwise separable convolution to reduce the number of parameters and computational operations, increasing efficiency. EfficientNet-B0 has a smaller scaling factor compared to other variants, resulting in a smaller model size and flexible, adjustable efficiency and speed, making it suitable for various application needs [31].

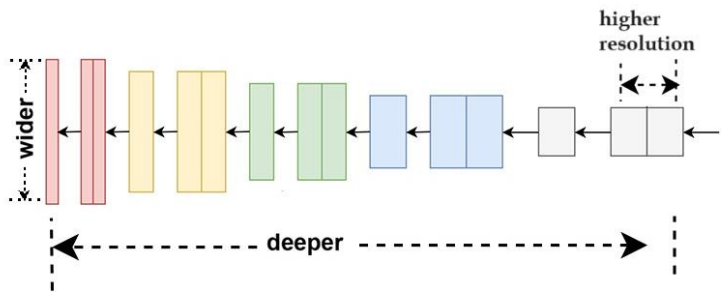

Figure 3. Compound Scaling model in EfficientNet [32]

MobileNetV2 is a lightweight deep neural network architecture designed to balance image resolution, network width, and depth for improved accuracy. It incorporates modifications to the network depth of MobileNetV2 to maintain gradient stability, reducing issues like gradient vanishing or exploding [33]. MobileNetV2's primary components include depthwise separable convolutions, bottlenecks, and inverted residuals. Depthwise separable convolutions are a combination of depthwise convolutions and pointwise convolutions. Depthwise convolutions apply a single filter to each input channel, while pointwise convolutions combine the outputs of depthwise convolutions using a 1x1 convolution. This approach significantly reduces the number of parameters and computational operations compared to standard convolutions.

Furthermore, MobileNetV2 introduces linear bottlenecks, a form of residual connection. These bottlenecks consist of a 1x1 convolution followed by a ReLU activation function, a 3x3 depthwise separable convolution, and another 1x1 convolution. Linear bottlenecks help to maintain representational power while reducing computational complexity. Inverted residuals are a modification of traditional residual connections. They reverse the order of the 1x1 and 3x3 convolutions in the bottleneck, allowing for a more efficient information flow.

Additionally, MobileNetV2 utilizes an improved Bottleneck module with a channel attention mechanism to extract more effective features from complex backgrounds [34]. MobileNetV2 is part of a broader landscape of efficient neural network architectures, including PeleeNet, DualConv, and Universally Slimmable Networks, all aimed at optimizing accuracy and efficiency trade-offs for various tasks like image classification, object detection, and semantic segmentation [35]. Figure 4 shows MobileNetV2-based architecture.

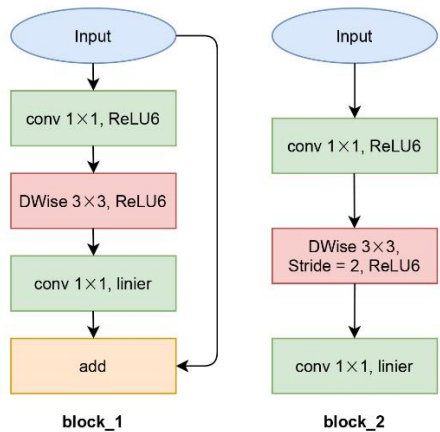

Figure 4 MobileNetV2 block

ShuffleNet is a lightweight convolutional neural network architecture designed for mobile devices and embedded systems. It proposed pointwise group convolutions and channel shuffling to reduce computational complexity and memory usage while maintaining accuracy significantly. ShuffleNet has two primary components, including Pointwise group convolutions and Channel shuffle. Pointwise group convolutions are 1x1 convolutions that divide input channels into groups and apply separate filters to each group. This approach reduces the number of parameters and computational operations compared to standard convolutions. After applying pointwise group convolutions, the channel shuffle operation rearranges the channels to facilitate information flow between groups. This approach helps to maintain representational power, minimize the number of parameters, and reduce the risk of overfitting [36].

Furthermore, non-linear activation functions, such as ReLU, are applied to introduce non-linearity into the model. Finally, the output data is produced after passing through the network. ShuffleNet's primary advantage is its ability to significantly reduce computational complexity and memory usage while maintaining accuracy. ShuffleNet has been proven to have notable performance in several cases of image classification, although on a lightweight model [37]. Figure 5 shows the ShuffleNet architecture block.

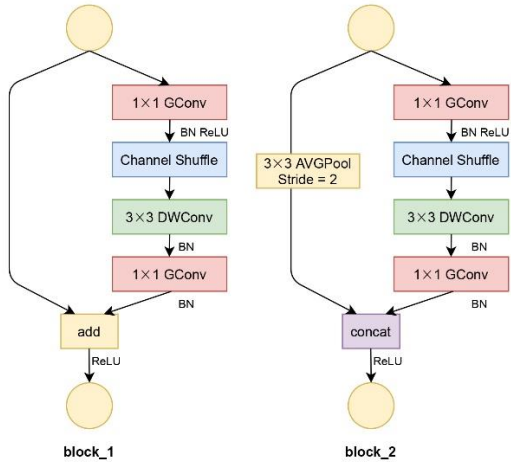

Figure 5. ShuffleNet main block

## Implementation Details

We implemented our CNN model using the Python version 3.8.5 and TensorFlow version 2.6.0 deep learning framework. The model was trained and tested on a workstation with an NVIDIA GeForce RTX 3090 GPU and 64 GB of RAM. During the experiment, we implemented various hyperparameters during the training processes. To obtain an optimal model, the learning rate was adjusted between $10^{-1}$ and $10^{-5}$. Specifically, two sets of experiments were conducted with different numbers of epochs, 200 and 300. These values were selected based on our preliminary analysis and prior knowledge of deep learning models. By varying the number of epochs, we aimed to achieve better generalization capabilities while avoiding potential issues such as overfitting or underfitting. To avoid overfitting caused by the use of highly similar images in our dataset, we adopted an effective regularization technique known as "early stopping." This method involves monitoring the loss function throughout the training process and terminating it prematurely when the validation loss does not decrease for a specified number of consecutive iterations (in our case, 15). Consequently, this approach prevents the model from excessively fitting to noise present in the data, leading to improved robustness and applicability.

Furthermore, we carefully configured several essential parameters. Firstly, a reasonable batch size of 32 was determined through empirical testing, balancing both computational efficiency and statistical accuracy. Secondly, we opted for the widely-used Adam optimizer, given its demonstrated effectiveness in numerous applications. As a stochastic gradient descent algorithm, Adam employs adaptive moment estimation to adjust weight updates dynamically, thereby facilitating faster convergence rates and enhanced overall performance compared to traditional methods. Lastly, regarding the selection of the appropriate loss function, we chose categorical cross entropy due to its suitability for multi-class classification problems.

## Evaluation Metric

Evaluation measures are crucial in assessing the performance of deep learning models. In this research, we will focus on four commonly used evaluation metrics: accuracy, recall, precision, and F-measure. Accuracy: Accuracy is a simple measure that calculates the proportion of correctly predicted instances over the total number of instances. It is computed as follows:

$$Accuracy = \frac{TP + TN}{TP + FP + TN + FN} \tag{1}$$

where TP (True Positives) are the instances correctly classified as positive, TN (True Negatives) are the instances correctly classified as negative, FP (False Positives) are the instances incorrectly classified as positive, and FN (False Negatives) are the instances incorrectly classified as negative.

Recall: Recall, also known as sensitivity or true positive rate, measures the proportion of true positive instances that the model correctly identifies. It is calculated as follows:

$$Recall = \frac{TP}{TP + FN} \tag{2}$$

Precision: Precision measures the proportion of true positive instances among all instances classified as positive by the model. It is computed as follows:

$$Precision = \frac{TP}{TP + FP} \tag{3}$$

F-Measure: The F-measure combines precision and recall into a single metric, providing a balanced evaluation of a model's performance. The harmonic mean of precision and recall is used to calculate the F-measure:

$$F - measure = \frac{2 \times Recall \times Precision}{Recall + Precision} \tag{4}$$

In this research, we will report the performance of our models using these evaluation measures. By analyzing the accuracy, recall, precision, and F-measure, we can better understand their strengths and weaknesses and make informed decisions about the best model for our application.

To assess the efficiency and evaluation parameters of the developed model, we used two additional metrics: learning parameters and energy usage. These metrics provide insights into both the model's learning effectiveness and its energy consumption profile. Learning parameters represent the number of parameters that should be used in the inference process. The energy usage represents the comparative energy consumption of each deep learning model rather than the absolute amount. The largest model is assigned a value of 1 or 100%, with other models reflecting their relative energy consumption compared to this benchmark.

## Result and discussion

In this section, we present and discuss the result, which investigated the best lightweight model for rice leaf disease detection and classification. Our approach combined three mobile-based pre-trained CNN models with dual dropout layer regularization. The dual dropout layer was applied to improve the model's ability to learn and prevent overfitting during the training process. We evaluate our approach on a public dataset and compare it to existing state-of-the-art methods. The results demonstrate the effectiveness of our approach, achieving higher accuracy and robustness than previous methods.

### Training result

Figure 6 shows the learning curve of the EfficientNet-B0 model using five different learning rate values, ranging from $10^{-1}$ to $10^{-5}$. The x-axis represents the number of epochs, which reflects the number of iterations during the training process. Meanwhile, the y-axis indicates the value of loss, which measures the degree of deviation between the predicted and actual values.

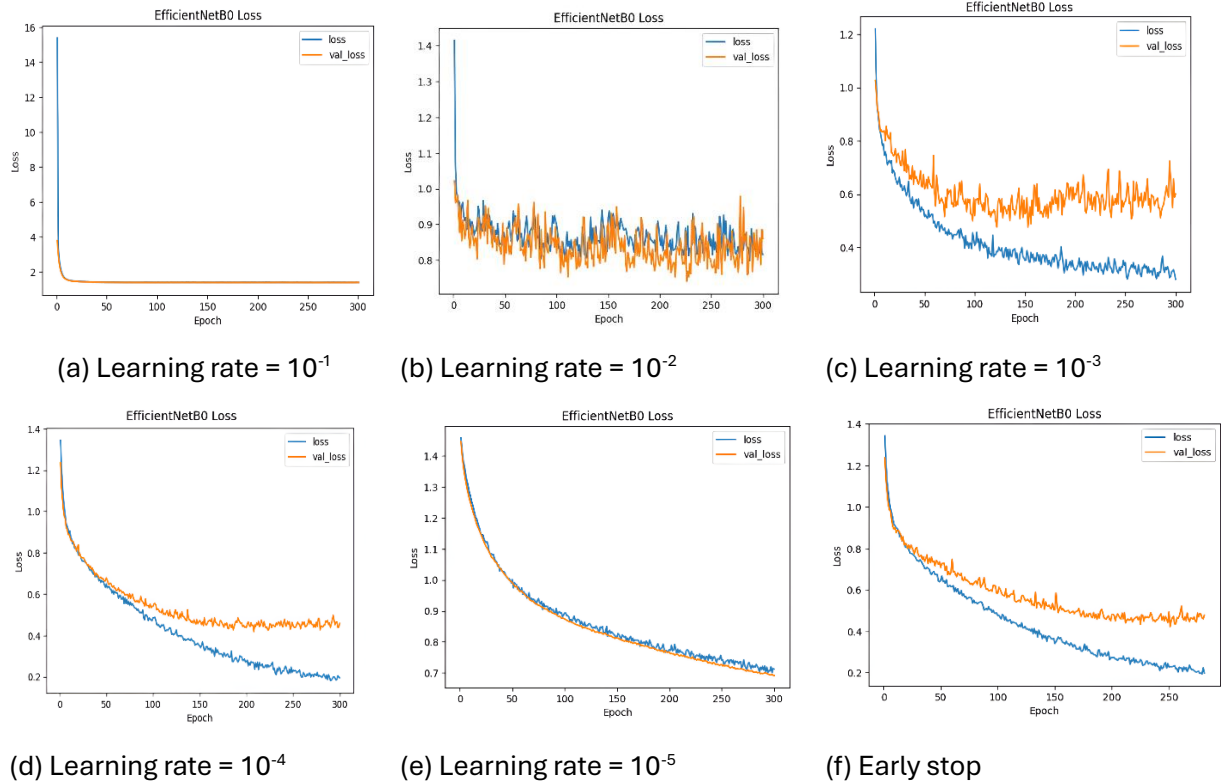

Figure 6 Training result for EfficientNet-B0 pre-trained model

Based on Figure 6(a), the curve shows a rapid decrease in loss value at the beginning of training, and the training loss appears to be close to the validation loss, indicating that the model has a potential for underfitting. Using a learning rate of $10^{-1}$ results in a relatively large loss value, indicating convergence at a loss of 2. Furthermore, Figure 6(b) depicts the loss function of an EfficientNet-B0 model trained with a learning rate of $10^{-2}$. In the graph, we can see that the training loss initially decreases as the number of epochs increases. However, the loss starts to fluctuate around 100 epochs, which could be a sign of overfitting. This performance could be due to a high learning rate. Although it shows better training performance than Figure 6(b), Figures 6(c) and 6(d) show the same trend, where the validation loss does not decrease around 100 epochs and even starts to increase after 200 epochs. Figure 6(e) shows that the curve with this learning rate has the slowest but most stable decrease in loss. However, the minimum loss value has yet to be reached after 300 epochs, indicating a very slow and inefficient training process. Due to unsatisfactory performance, we applied early stopping on the EfficientNet-B0 model that trained using $10^{-4}$ to prevent overfitting in the model. As shown in Figure 6(f), due to the early stopping criteria, the training process stopped at 285 epochs.

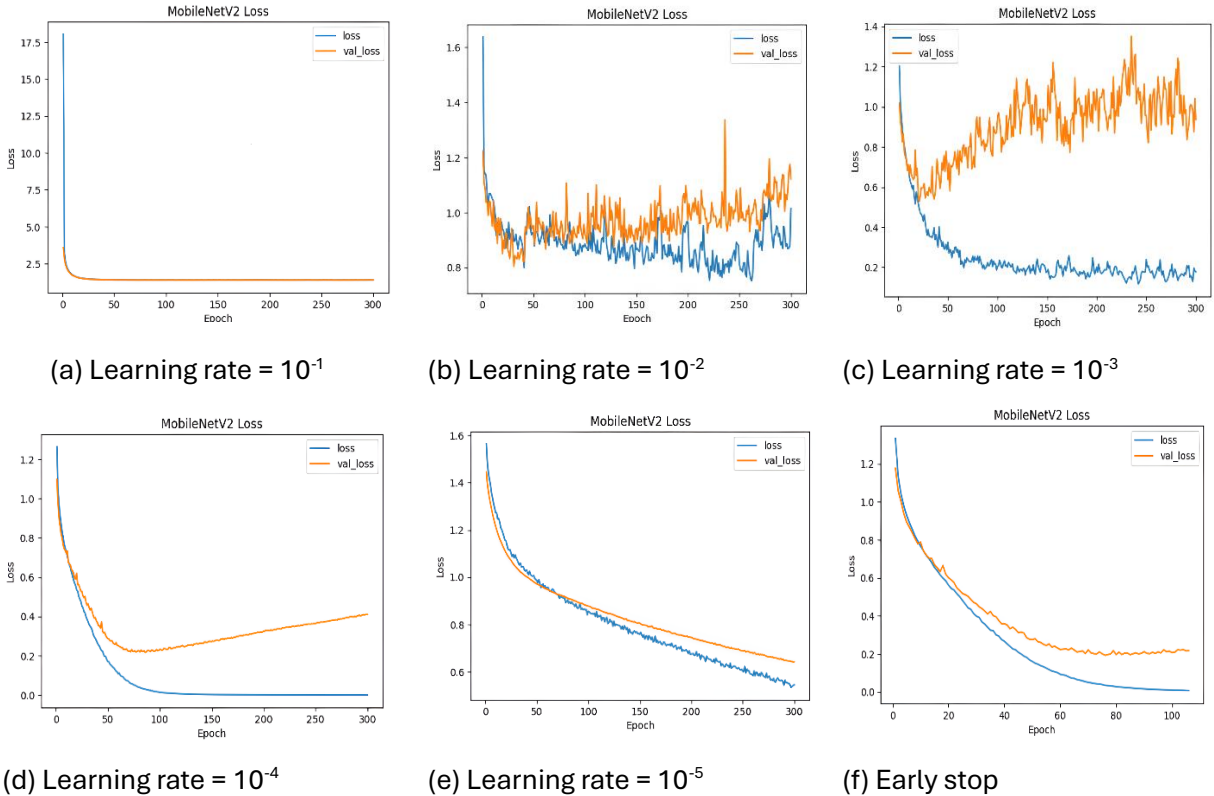

Figure 7 Training result for MobileNetV2 pre-trained model

Figure 7 shows the loss function of a MobileNetV2 model trained with different learning rates. The x-axis represents the number of epochs, which corresponds to the number of times the training data has been iterated through. The y-axis represents the loss, which is a numerical value indicating how well the model is performing on the training task. A lower loss signifies a better performance. Figure 7(a) shows the learning curve of the MobileNetV2 that was trained using a learning rate of $10^{-1}$. As shown in Figure 7(a), the curve illustrates a rapid decrease of loss value at the beginning of training, but the loss value convergence on loss value of 2. It is indicated that the model failed to learn the pattern from the dataset to generalize a model.

Figure 7(b) depicts the loss function of a MobileNetV2 model trained with a learning rate of $10^{-2}$. In the graph, we can see that the training loss initially decreases as the number of epochs increases. However, the loss starts to fluctuate around 100 epochs, and the validation loss starts to increase after 200 epochs, which could be a sign of overfitting caused by a high learning rate. Figures 7(c) and 7(d) depict the MobileNetV2 model that is tranined with learning rates $10^{-3}$ and $10^{-4}$. Based on Figures 7(c) and 7(d), the training loss initially decreases, but the validation loss starts to increase around 50, epochs which could be a sign of overfitting. In addition, Figure 7(c) has a more fluctuated learning curve compared with Figure 7(d). Figure 7(e) shows the curve with the slowest but most stable decrease in loss. However, the minimum loss value has not been achieved after 300 epochs, indicating a very slow and inefficient training process. Similar to the EfficientNet-B0 model, we applied early stopping on the EfficientNet-B0 model that trained using $10^{-4}$ to prevent overfitting in the model because of unsatisfactory performance. As shown in Figure 7(f), due to the early stopping criteria, the training process stopped at 112 epochs.

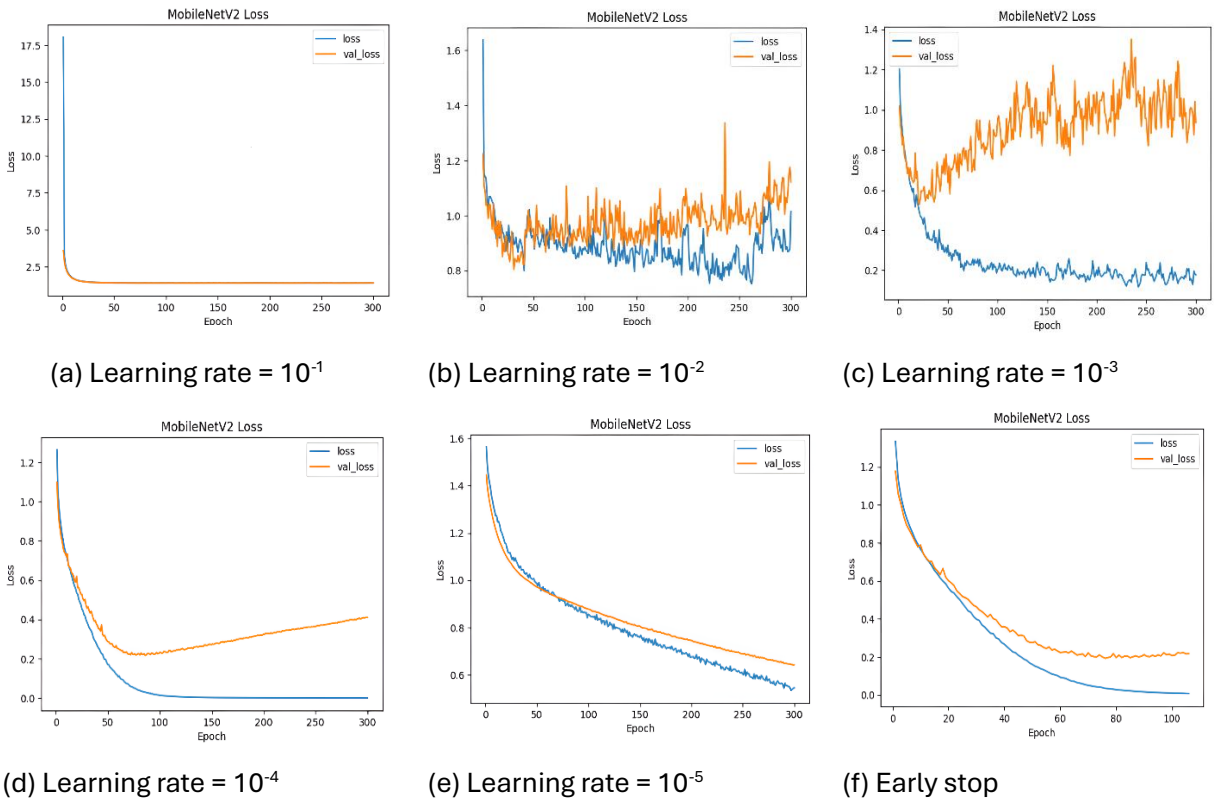

(a) Learning rate = $10^{-1}$  (b) Learning rate = $10^{-2}$  (c) Learning rate = $10^{-3}$

(d) Learning rate = $10^{-4}$  (e) Learning rate = $10^{-5}$  (f) Early stop

Figure 8 Training result for ShuffleNet pre-trained model

Figure 8 presents the learning curve of a model trained on ShuffleNet with different learning rates. Based on Figures 8(a) and 8(b), the learning curve indicates overfitting, even before the model has been trained up to epoch 50. Figure 8(c) is more stable than Figures 8a and 8b, but both models have a high loss value, which is still above 0.5. The model trained with a learning rate of $10^{-3}$ even has the potential for overfitting, as there is no reduction in validation loss after epoch 250. Figures 8(d) and 8(e) show that the curve with this learning rate has the slowest but most stable decrease in loss. However, the minimum loss value has yet to be reached convergency after 300 epochs, indicating a very slow and inefficient training process. Due to the unsatisfactory performance, we applied early stopping to the model with a learning rate of $10^{-3}$, as this model is more stable, even if it is slightly less efficient. Figure 8(f) shows the model's training process after implementing early stopping criteria.

## Testing Result

Table 1 presents the performance of three mobile-based pre-trained models: EfficientNet-B0, MobileNetV2, and ShuffleNet. Those models were evaluated over 300 epochs and with early stopping. The key performance metrics assessed were Accuracy, Recall, Precision, and F-Measure, along with the number of learning parameters for each model.

Table 1. Testing result of the three lightweight CNN pre-trained model

| Pre-trained Model | Training Epoch | Accuracy | Recall | Precision | F-Measure | Learning parameters |
|---|---|---|---|---|---|---|
| EfficientNet-B0 | 300 | 99.52% | 99.59% | 99.58% | 99.58% | 5.4 millions |
|  | Early stop | 99.52% | 99.55% | 99.50% | 99.52% |  |
| MobileNetV2 | 300 | 84.21% | 84.23% | 84.35% | 84.13% | 3.5 millions |
|  | Early stop | 79.90% | 78.83% | 78.85% | 78.83% |  |
| ShuffleNet | 300 | 66.03% | 65.64% | 65.78% | 65.56% | 1.4 millions |
|  | Early stop | 66.51% | 67.68% | 67.60% | 67.19% |  |

EfficientNet-B0 demonstrated exceptional performance, with accuracy, recall, precision, and F-measure values all exceeding 99.5%. It indicates the model's robustness and reliability in accurately identifying both positive and negative cases. The marginal differences between metrics after early stopping suggest that EfficientNet-B0 maintains its performance even with reduced training time. MobileNetV2 showed moderate performance with an accuracy of 84.21% over 300 epochs. The balance between recall (84.23%) and precision (84.35%) led to a consistent F-measure (84.13%). However, early stopping significantly reduced the performance, with accuracy dropping to 79.90%, indicating that MobileNetV2 relies heavily on full training epochs to achieve optimal results.

Furthermore, ShuffleNet had the lowest performance, with an accuracy of 66.03%. Despite this, early stopping resulted in a slight improvement across all metrics, with accuracy increasing to 66.51%. It suggests that ShuffleNet might be overfitting with extended training and benefits from early stopping to avoid this issue.

EfficientNet-B0, with 5.4 million learning parameters, correlates its high parameter count with superior performance. It indicates that the model's complexity and capacity contribute to its ability to learn and generalize effectively. MobileNetV2 has 3.5 million parameters, striking a balance between model size and performance. However, its sensitivity to training duration highlights the need for adequate epochs to leverage its architecture fully. ShuffleNet, the model with the smallest parameter count (1.4 million), provides the lowest performance among the three. However, its lightweight nature makes it suitable for scenarios with limited computational resources, and its performance can be slightly improved with early stopping.

EfficientNet-B0 exhibited minimal impact from early stopping, maintaining high performance across all metrics. It suggests the model's robustness and ability to achieve near-optimal performance without requiring the full duration. MobileNetV2, on the other hand, experienced a significant drop in performance with early stopping, indicating its dependency on extended training epochs. It highlights the need for careful consideration of training duration to avoid underfitting. ShuffleNet showed an improvement in performance metrics with early stopping, suggesting that it might suffer from overfitting during prolonged training. Early stopping can thus be beneficial for ShuffleNet, helping to achieve a better balance between training duration and performance.

Based on the results, EfficientNet-B0 stands out as the best-performing model across all metrics, demonstrating its advanced architecture's efficacy and the advantages of a higher parameter count. MobileNetV2 shows moderate performance but is significantly impacted by early stopping, indicating a need for full training epochs. ShuffleNet, while having the lowest performance, benefits from early stopping, making it suitable for resource-constrained environments where computational efficiency is a priority. The choice of model should be based on the specific requirements and constraints of the application, balancing performance needs with available computational resources.

## Comparison with others

Table 2 provides a summary of the performance metrics for seven pre-trained models: VGG-16 [38], XceptionNet [39], ResNet50 [40], GhostNet, EfficientNet-B0, MobileNetV2, ModLeafNet, ShuffleNet. The key metrics evaluated are Accuracy, the number of learning parameters for each model, and energy usage. The energy usage represents the comparative energy consumption of each deep learning model rather than the absolute amount. The highest learning parameter is assigned a value of 1.0, with other models reflecting their relative energy consumption compared to this benchmark.

Table 2 Comparison performance with previous studies

| Pre-trained Model | Accuracy | Learning parameters | Energy used |
|---|---|---|---|
| VGG16 [38] | 91.66% | 138 millions | 1.00 |
| Xception [39] | 71. 95% | 27.3 millions | 0.20 |
| ResNet50 [40[ | 91.90% | 25 millions | 0.18 |
| GhostNet [41] | 89.30% | 5.2 millions | 0.04 |
| EfficientNet-B0 | 99.52% | 5.4 millions | 0.04 |
| MobileNetV2 | 84.21% | 3.5 millions | 0.03 |
| ModLeafNet [39] | 87.76% | 1.3 millions | 0.01 |
| ShuffleNet | 66.03% | 1.4 millions | 0.01 |

EfficientNet-B0 achieved a high accuracy of 99.52% with a moderate parameter count of 5.4 million. This result indicates that EfficientNet-B0 is highly efficient, providing excellent performance without excessive computational cost. This balance of high accuracy and moderate complexity makes EfficientNet-B0 suitable for various applications, including those requiring a combination of accuracy and efficiency. Furthermore, GhostNet recorded an accuracy of 89.30% with 5.2 million parameters. While less high-performing than EfficientNet-B0, GhostNet still offers a respectable accuracy with a similar parameter count. This model strikes a good balance between performance and computational efficiency, making it suitable for scenarios where moderate accuracy is acceptable and computational resources are limited. VGG16 achieved an accuracy of 91.66% with a significantly high parameter count of 138 million. This model's high accuracy comes at the cost of substantial computational resources, making it less suitable for environments with limited computational power. VGG16 is more appropriate for applications where the highest accuracy is critical and there is ample computational capacity to handle the large model size.

ResNet50 achieved an outstanding accuracy of 91.90% with 25 million parameters. This high level of performance with moderate parameter counts highlights ResNet50's efficiency and suitability for applications requiring high accuracy without the need for an excessively large model. Modified LeafNet (ModLeafNet) recorded a high accuracy of 87.76% with a low parameter count of 1. 3 million parameters, the lowest among others. While offering good performance, ModLeafNet's small number of parameters makes it highly suitable for highly resource-constrained environments where computational efficiency is paramount and performance requirements are minimal.

MobileNetV2 demonstrated an accuracy of 84.21% with 3.5 million parameters, reflecting moderate performance. The lower parameter count highlights its design for resource-constrained environments, such as mobile and embedded applications, where computational efficiency is prioritized over achieving the highest possible accuracy. ShuffleNet, with the lowest accuracy of 66.03% and a small parameter count of 1.4 million, is designed for extreme efficiency and less computational cost. Similar to ModLeafNet, despite its low accuracy, ShuffleNet is highly suitable for highly resource-constrained environments.

EfficientNet-B0 stands out for its high accuracy with moderate parameter counts, indicating an optimal balance between performance and computational efficiency. VGG16 and XceptionNet, while highly accurate, are computationally expensive due to their large number of parameters, making them less suitable for resource-constrained environments. For applications with severe computational constraints, ModLeafNet, MobileNetV2, and ShuffleNet are preferable despite their lower accuracy. GhostNet offers a middle ground with decent accuracy and a moderate parameter count, which is suitable for applications requiring a balance between performance and efficiency.

## Conclusions

This research constructed a classification model for rice leaf disease using pre-trained CNN models. The results showed that the EfficientNet-B0 model, trained with an early stop and a learning rate of $10^{-4}$, was the best model with an accuracy of 99.52%, precision of 99.59%, recall of 99.58%, and F1-score of 99.58%. This performance was better than MobileNetV2, which achieved an accuracy of 84.21%, and ShuffleNet, which achieved an accuracy of 66.03%. EfficientNet-B0 has an advantage in improving accuracy compared to MobileNet-V2 and ShuffleNet architectures. Based on the results, the EfficientNet-B0 model can be used for mobile devices due to its relatively small model size, with only 5.4 M learning parameters. Developing smaller and more efficient models is essential for reducing the environmental impact of AI systems. Smaller models require fewer computational resources, leading to lower energy consumption and reduced greenhouse gas emissions. Therefore, using EfficientNet-B0 for plant disease classification in rice plants is a step towards creating more sustainable and environmentally friendly AI systems. Future work will deploy smaller models that can handle more actual classes, enabling them to manage all types of leaf diseases in rice plants.